\documentclass{article}

\PassOptionsToPackage{numbers, compress}{natbib}
\usepackage[preprint]{neurips_2020}

\usepackage[utf8]{inputenc} 
\usepackage[T1]{fontenc}    
\usepackage{hyperref}       
\usepackage{url}            
\usepackage{booktabs}       
\usepackage{amsfonts}       
\usepackage{nicefrac}       
\usepackage{microtype}      
\usepackage[pdftex]{graphicx}

\usepackage{wrapfig}
\usepackage{amsmath}

\title{Emergence of Spatial Coordinates via Exploration}

\author{
  Alban Laflaqui\`ere\\
  AI Lab, SoftBank Robotics Europe\\
  43 Rue du Colonel Pierre Avia, 75015 Paris, France\\
  \texttt{alaflaquiere@softbankrobotics.com}\\
}

\begin{document}

\maketitle

\begin{abstract}
Spatial knowledge is a fundamental building block for the development of advanced perceptive and cognitive abilities. Traditionally, in robotics, the Euclidean $(x,y,z)$ coordinate system and the agent's forward model are defined a priori. We show that a naive agent can autonomously build an internal coordinate system, with the same dimension and metric regularity as the external space, simply by learning to predict the outcome of sensorimotor transitions in a self-supervised way.  
\end{abstract}

\section{Introduction}
\label{sec:introduction}

Human babies gradually make sense of their environment through their experiences.
Via a continuous active engagement, they develop progressively richer ways to interact with the world and richer concepts to think about it.
Understanding the underlying mechanisms of this development is of prime importance if we want to build robots that display comparable capacities for autonomy and, eventually, human-level intelligence.
\\
Because humans are the product of millions of years of evolution, babies are not born as blank slates. Some prior knowledge about the world and themselves is hard-coded in their body and brain, shaping their physical and cognitive development.
However, it is an open question how much of this prior knowledge is required to bootstrap the learning process.
This inquiry is especially relevant for roboticists who have to act the part of evolution for their system.
\\
This also raises the question of scaffolding: what are the fundamental building blocks on top of which more knowledge can be built?
In this work, we suggest that \emph{space} is such a building block.
By \emph{space} we mean the knowledge for the agent that it is moving in an external Euclidean space.
In robotics, such knowledge is traditionally hard-coded a priori as a predefined $(x,y,z)$ coordinate system coupled with a forward model mapping a motor state to such a position.
Yet, could a naive agent autonomously acquire an equivalent spatial insight by interacting with its environment?
This sounds like a challenge.
Firstly, this means building a stable spatial representation when the sensorimotor experiences collected by the agent vary greatly as it explores the world.
Secondly, this means discovering a Euclidean structure in a sensorimotor manifold that displays a very different structure.
\\
Nonetheless, Poincar\'e theorized more than a century ago that it should be possible to do so without any a priori knowledge about the sensorimotor mapping \cite{Poincare1895}.
Recently, this spurred a field of research dedicated to materialize his intuition by studying how spatial knowledge can be grounded in uninterpreted sensorimotor experiences \cite{philipona2003there, Laflaquiere2012, Laflaquiere2013, Laflaquiere2015a, marcel2015building, terekhov2016space, marcel2016building, laflaquiere2018discovering, ortiz2018learning}.
This paper extends this line of work by studying how sensorimotor transitions contain information about the Euclidean structure of the external space, and how this structure is naturally captured when doing sensorimotor prediction.
In particular, it further develops the experiments of \cite{laflaquiere2019unsupervised} by considering a mobile robot (instead of a static one) exploring multiple environments (instead of one), as well as the impact of extraneous motors.

\section{Problem formulation}
\label{sec:problem}

At time $t$, let's consider a naive agent that has access to uninterpreted sensorimotor states $(\mathbf{m}_t, \mathbf{s}_t)$, where $\mathbf{m}_t$ corresponds to its body proprioception and $\mathbf{s}_t$ corresponds to the exteroception produced by a sensor somewhere on its body.
Like most animals, we assume that this agent is mobile and can freely generate the same motor state $\mathbf{m}_t$ independently of its position.
This is comparable to the way we can walk between two positions in a room and reproduce the same postures in both positions.
Moreover, let's assume that the agent is in an environment whose overall state is denoted $\epsilon$. It encapsulates every properties of the environment that the agent can have access to (for example, the (shape, color, position...) of all objects around it).
Putting aside potential sensorimotor noise, the sensory state $\mathbf{s}_t^{\epsilon}$ is then defined via an (unknown) sensorimotor mapping $\mathbf{s}_t^{\epsilon} = \phi_\epsilon(\mathbf{m}_t)$, parametrized by $\epsilon$.
Note that the agent, being naive, does not have access to the parameter $\epsilon$ and can only process raw $(\mathbf{m}_t, \mathbf{s}_t)$ states.
\\
On the other hand, the agent is also embedded in space, and changing its motor state changes the external position of its sensor.
We define this position, denoted $\mathbf{p}_t$, in a typical Euclidean frame of reference centered on the base of the body, i.e. the part of the body that stays in contact with the environment (the ground) while the sensor moves.
We can then define a "sensory-positional" counterpart to the previous sensorimotor mapping: $\mathbf{s}_t^{\epsilon} = \chi_\epsilon(\mathbf{p}_t)$, where $\mathbf{p}_t = f(\mathbf{m}_t)$ is the forward model associated with the sensor position.
Once again, note that the agent doe not have direct access to this egocentric position $\mathbf{p}_t$, nor to $\chi_\epsilon$ or $f$.

From raw $(\mathbf{m}_t, \mathbf{s}_t)$ states, how can the agent build an internal representation $\mathbf{h}_t$ with the same geometric structure as the actual sensor's egocentric position $\mathbf{p}_t$?
This is not straightforward.
The motor and sensory spaces in which $\mathbf{m}_t$ and $\mathbf{s}_t$ respectively live have a different dimension than the external Euclidean space of $\mathbf{p}_t$.
They also have a different metric, such that two displacements $\overrightarrow{\mathbf{p}_t \mathbf{p}_{t+1}}$ of equal amplitude a priori correspond to two motor changes $\overrightarrow{\mathbf{m}_t \mathbf{m}_{t+1}}$ of different amplitudes, as well as two sensory changes $\overrightarrow{\mathbf{s}_t \mathbf{s}_{t+1}}$ of different amplitudes. The situation is even worse in the sensory case as $\mathbf{s}_t$, and thus the change $\overrightarrow{\mathbf{s}_t \mathbf{s}_{t+1}}$, vary with the (unknown) state of the environment $\epsilon$.
This sensory instability is particularly problematic, as spatial information has to necessarily be extracted from the sensory flown which is the only one carrying information about the external world.

Fortunately, despite its "blooming buzzing confusion" \cite{james2007principles}, the sensorimotor flow contains an underlying spatial structure.
Let's then build up some intuitions. 
When exploring any environment, motor states from the equivalence class $[\mathbf{m}]_f = \{\mathbf{m}'|f(\mathbf{m'}) = f(\mathbf{m})\}$ associated with the same sensor position $\mathbf{p}$ generate equivalent sensory states: $\forall \epsilon, \phi_\epsilon([\mathbf{m}]_f) = \chi_\epsilon(\mathbf{p}) = \mathbf{s}^\epsilon$.
When predicting the sensory outcome of a transition $(\mathbf{m}_t, \mathbf{s}_t, \mathbf{m}_{t+1})\rightarrow \tilde{\mathbf{s}}_{t+1}$ in any environment $\epsilon$, the agent can learn that all motor states $[\mathbf{m}_{t+1}]_f$ lead to the same $\mathbf{s}_{t+1}$ and are thus equivalent.
\\
A similar, but more sophisticated, reasoning also explains how the agent can discover the metric equivalence between different motor changes $\overrightarrow{\mathbf{m}_t \mathbf{m}_{t+1}}$.
In this case we need to consider what happens when the agent moves its base during the exploration of an environment.
First, as wisely observed by Poincar\'e, moving the base with respect to the environment is equivalent to an opposite move of the environment with respect to the base, or 
$\chi_{\epsilon}(\mathbf{p} + \delta) = \chi_{\epsilon - \delta}(\mathbf{p})$ (this is sensory compensability \cite{Poincare1895}).
We can thus denote a base change as an environmental change $\epsilon \rightarrow \epsilon'$ (see \cite{laflaquiere2019unsupervised}).
During such base displacements, the sensory manifold that the agent can experience is unchanged but shifted (rigidly) with respect to the space of positions $\mathbf{p}$.
Likewise it is shifted with respect to the motor space, but this time non-rigidly because of the non-linear forward model $f$.
This means that the equivalence class
\[
\big[(\mathbf{m}_a,\mathbf{m}_b)\big]_{ff} =
 \big\{ (\mathbf{m}_{a'}, \mathbf{m}_{b'}) | \overrightarrow{f(\mathbf{m}_{a'})f(\mathbf{m}_{b'})} = \overrightarrow{f(\mathbf{m}_a)f(\mathbf{m}_b)} \big\}
\]
associated with the same external displacement $\overrightarrow{\mathbf{p}_a\mathbf{p}_b}$ generate equivalent sensory changes (see App.~\ref{sec:appendix}):
\[
\forall \epsilon,
\forall (\mathbf{m}_{a'},\mathbf{m}_{b'}) \in \big[(\mathbf{m}_a,\mathbf{m}_b)\big]_{ff},
\forall \epsilon' : \phi_{\epsilon'}(\mathbf{m}_{a'})= \mathbf{s}_a^{\epsilon} \Rightarrow
\phi_{\epsilon'}(\mathbf{m}_{b'})= \mathbf{s}_b^{\epsilon}.
\]
More intuitively this means, for example, that moving the sensor from the bottom to the top of an object engages different motor transitions depending on whether the object is in front of the agent or on its right.
Yet, the size of the object is constant and thus the actual sensor displacement associated with these different motor changes are identical.
When predicting the sensory outcome of a transition $(\mathbf{m}_t, \mathbf{s}_t, \mathbf{m}_{t+1})\rightarrow \tilde{\mathbf{s}}_{t+1}$ after moving the base in any environment $\epsilon$, the agent can learn that all motor changes in $\big[(\mathbf{m}_t,\mathbf{m}_{t+1})\big]_{ff}$ are equivalent.

Interestingly, the equivalence relations we described are independent of the content of the environment $\epsilon$, and hold throughout the agent's lifetime.
Moreover we hypothesize that an agent trying to predict its own sensorimotor experience would benefit from capturing them, as knowing that different motor changes are equivalent would facilitate the predictive task.

In the following sections, we experimentally test this hypothesis by having an agent learn to predict its own sensorimotor experience with a simple neural network, and compare the internal motor representation it learns to the ground-truth sensor position it does not have access to.

\section{Experiment}
\label{sec:experiment}

\begin{wrapfigure}{R}{30mm}
\centering
\includegraphics[width=30mm]{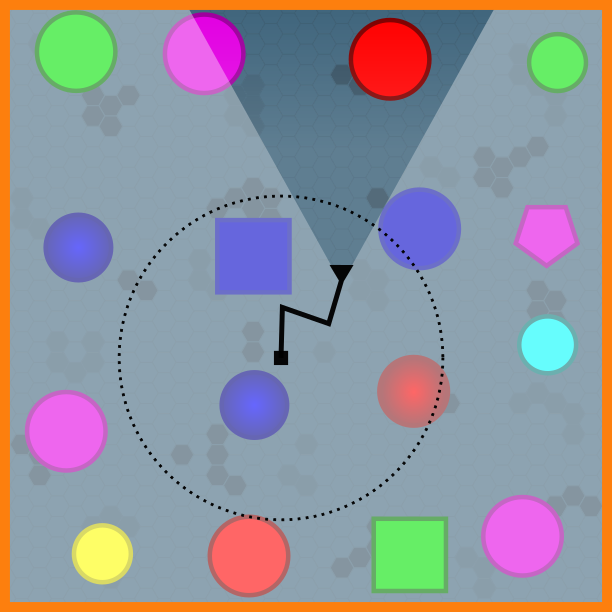}
\caption{Simulation}
\label{fig:simulation}
\end{wrapfigure}

We simulate an agent composed of a mobile base and a 3-segments planar arm equipped with a camera on its end-effector (see Fig.~\ref{fig:simulation}).
It can move its whole body by hopping its base around, and can freely generate arm postures represented by motor states $\mathbf{m}_t \in \mathbb{R}^4$ (proprioceptive readings).
One of the motor dimensions ($m_4$) is added as a distractor and does not influence the actual arm posture.
In the current simulation, the orientation of the camera is kept constant (with respect to the base), so that the sensor only experiences translations during the exploration.
This way, its egocentric position is characterized by two coordinates $\mathbf{p}_t = (x_t, y_t)$ only.
One could in theory consider additional translations along $z$ and rotations of the sensor, but this would make the visualization of the motor representation more intricate, especially in such a short paper.
The camera itself provides an instantaneous sensory state $\mathbf{s}_t \in \mathbb{R}^{768}$ corresponding to a $16 \times 16$ rgb image flattened as a raw vector in which the array structure has been discarded.

The agent is placed in $10$ successive 3D environments that it can explore randomly.
Each environment is a square room regularly filled with random objects\footnote{The arm is placed high enough in the room to avoid collisions with objects.} (see Fig.~\ref{fig:simulation}).
We compare three modes of exploration to test our hypothesis.
In the \textbf{nominal mode}, the agent collects sensorimotor transitions $(\mathbf{m}_t, \mathbf{s}_{t}, \mathbf{m}_{t+1}) \rightarrow \mathbf{s}_{t+1}$ by repeatedly i) hopping randomly in the room (moving its base) and then ii) experiencing two successive random motor states $(\mathbf{m}_t, \mathbf{m}_{t+1})$ for which it receives two sensory states $(\mathbf{s}_t, \mathbf{s}_{t+1})$.
We compare it to two baselines designed to prevent the agent from experiencing the equivalence relations presented in Sec.\ref{sec:problem}.
In the first (\textbf{dynamic mode}), the agent also hops between $\mathbf{m}_t$ and $\mathbf{m}_{t+1}$, preventing it from experiencing transitions with a constant $\epsilon$.
It should thus not be able to discover the equivalent classes $[\mathbf{m}]_{f}$ and $\big[(\mathbf{m}_a,\mathbf{m}_b)\big]_{ff}$.
On the contrary, in the second (\textbf{static mode}), the agent never hops around and its base stay static in each environment, preventing it from experiencing transitions with different $\epsilon + \delta$.
This way, it cannot discover the equivalence class $\big[(\mathbf{m}_a,\mathbf{m}_b)\big]_{ff}$.
Regardless of the exploration mode, $10000$ transitions $(\mathbf{m}_t, \mathbf{s}_{t}, \mathbf{m}_{t+1}) \rightarrow \mathbf{s}_{t+1}$ are generated in each environment and used, as a whole, as training dataset for the neural network.

The neural network used for prediction chains up two MLPs.
The first one, of size $([4],$ $150,$ $100,$ $50,$ $[3])$, maps a motor state $\mathbf{m}_{t}$ to a motor representation $\mathbf{h}_{t}$ of dimension $3$.
The second, of size $([774],$ $200,$ $150,$ $100,$ $[768])$, takes as input the concatenation of two motor representations $(\mathbf{h}_{t}, \mathbf{h}_{t+1})$ generated by the first MLP and the sensory state $\mathbf{s}_{t}$, and outputs the sensory prediction $\tilde{\mathbf{s}}_{t+1}$.
It is trained with a standard MSE loss and Adam optimizer, for $50$ epochs and with a batch size of $128$.

\section{Results}
\label{sec:results}

We compare the motor representation $\mathbf{h}$ learned by the network to the ground-truth $\mathbf{p}$ we have access to using the following dissimilarity measure:
\[
D_{\alpha} = \frac{2}{N^2 - N}
\sum_{i=1}^N \sum_{j=i+1}^N
\big\vert d(\mathbf{h}_{i}, \mathbf{h}_{j}) - d(\mathbf{p}_{i}, \mathbf{p}_{j}) \big\vert
.
{e}^{-\alpha d(\mathbf{p}_{i}, \mathbf{p}_{j})}
\: \textrm{, with } \:
d(\mathbf{u}_i, \mathbf{u}_j) = \frac{\Vert \mathbf{u}_i - \mathbf{u}_j \Vert}{\max_{kl} \Vert \mathbf{u}_{k} - \mathbf{u}_{l} \Vert},
\]
with $\{ \mathbf{h}_i \}_{i=1}^N$ a set of representations collected by sampling regularly the motor space and feeding these samples to the first MLP, and $\{ \mathbf{p}_i \}_{i=1}^N$ the corresponding set of sensor positions.
Note that in practice, the manifold underlying $\mathbf{h}_i$ has no reason to be commensurate with the space of positions $\mathbf{p}_i$. We thus perform a linear regression of $\{ \mathbf{h}_i \}_{i=1}^N$ on $\{ \mathbf{p}_i \}_{i=1}^N$ and scale the positions with the resulting projection matrix $A . \mathbf{p}_i \rightarrow \mathbf{p}_i$ before calculating $D_\alpha$ (intercepts do not affect pairwise distances).
\\
Because the manifold on which the $\{ \mathbf{h}_i \}_{i=1}^N$ live is not necessarily flat, it is non-trivial to estimate its intrinsic dimension.
Instead we use $D_{10}$ as proxy for this measure.
Indeed if the distances $d(\mathbf{h}_{i}, \mathbf{h}_{j})$ are \emph{locally} commensurate with $d(\mathbf{p}_{i}, \mathbf{p}_{j})$ (small $D_{10}$), then their topology and intrinsic dimension can be considered similar.
Similarly, we use $D_0$ to estimate the general dissimilarity between the metrics of $\{ \mathbf{h}_i \}_{i=1}^N$ and $\{ \mathbf{p}_i \}_{i=1}^N$. This measure tends to $D_0 = 0$ when $\{ \mathbf{h}_i \}_{i=1}^N$ is flat and with the same pairwise distance ratios as $\{ \mathbf{p}_i \}_{i=1}^N$.

\begin{figure}[t]
\centering
\includegraphics[width=1\linewidth]{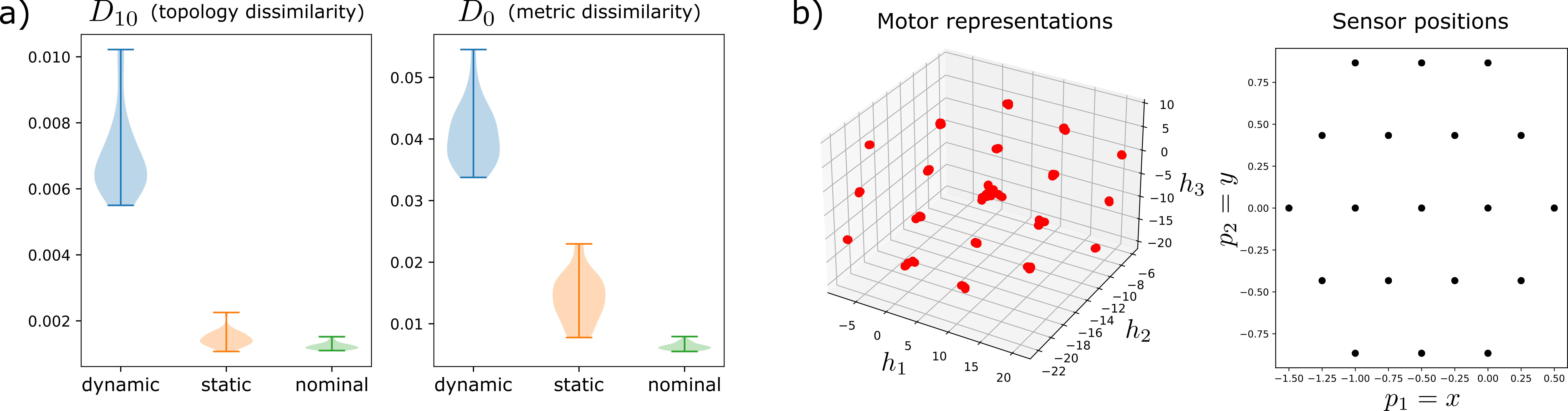}
\caption{a) Dissimilarity measures. b) Motor representation compared to ground-truth positions.}
\label{fig:stats}
\end{figure}

Figure~\ref{fig:stats} (a) shows the statistics for these two measures over 30 independent runs, and for the three modes of exploration.
We see that both dissimilarities are high in the dynamic mode. This was expected, as the agent can never experience any consistent transitions in which $\epsilon$ is stable. It thus cannot discover the equivalence classes.
\\
On the contrary, in the static mode the ``topological'' dissimilarity $D_{10}$ is low, while the ``metric'' dissimilarity $D_0$ displays some intermediary values (and a high variance).
Once again, this result was expected as the agent can discover the equivalence class $[\mathbf{m}]_{f}$.
It effectively discovers that each motor state $\mathbf{m}$ is associated with a 2D manifold of equivalent states (the 4D motor state has 2 redundant dimensions when setting the sensor position $(x, y)$) and represents them with the same $\mathbf{h}$.
The resulting manifold of representations is thus intrinsically of dimension $4 - 2 = 2$, like the space of positions $\mathbf{p}$.
However, because the base never moves, the agent cannot discover the equivalent class $\big[(\mathbf{m}_a,\mathbf{m}_b)\big]_{ff}$. The overall metric of $\{ \mathbf{h}_i \}_{i=1}^N$ thus differs from the one of $\{ \mathbf{p}_i \}_{i=1}^N$\footnote{Many $\{ \mathbf{h}_i \}_{i=1}^N$ correspond to identical positions, which explain why $D_{0}$ also decreases when $D_{10}$ does.}.
\\
Finally, in the nominal mode of exploration, $D_{10}$ is as low as in the static mode, and $D_{0}$ is the lowest of the three modes.
This shows that the overall metric of the motor representation $\{ \mathbf{h}_i \}_{i=1}^N$ is commensurate with the one of $\{ \mathbf{p}_i \}_{i=1}^N$ (the topology/dimension is necessarily similar when the metrics are similar).
The final motor representation learned by the network via sensorimotor prediction is thus a good proxy for the actual sensor coordinates $(x,y)$ in the external Euclidean space.
Figure \ref{fig:stats} (b) shows the $\{ \mathbf{h}_i \}_{i=1}^N$ obtained in one experiment and compares it to the ground-truth positions $\{ \mathbf{p}_i \}_{i=1}^N$.

Additional exploratory work, not presented in this short paper, indicate that these results are qualitatively robust to greater dimensions of $\mathbf{h}$ and more complex forward models (body with more motors and degrees of freedom of different natures).

\section{Conclusion}
\label{sec:conclusion}

We proposed that spatial knowledge is a fundamental building block for the development of perceptive and cognitive capacities, and wondered if such a knowledge can be acquired autonomously by a naive agent.
We theoretically studied how spatial structure induces structure in an agent's sensorimotor experience in the form of equivalence classes, and hypothesized that capturing such equivalence classes would be beneficial for an agent trying to predict its own sensorimotor experience.
Finally, we experimentally showed that such an agent can indeed capture this structure and autonomously build an internal representation that is geometrically similar to the external egocentric spatial configuration of its sensor.
Interestingly, this structure is independent of the content of the environment ($\epsilon$); a stability which makes it a promising building block for the accumulation of more sophisticated spatial knowledge throughout the life of the agent.

\bibliographystyle{unsrtnat}
\bibliography{biblio}

\newpage

\appendix
\section{Metric equivalence class}
\label{sec:appendix}

We assume here that $\chi$ is injective:
\begin{equation}
\forall \epsilon, \forall \mathbf{p}_u, \forall \mathbf{p}_v :
\chi_\epsilon(\mathbf{p}_u) = \chi_\epsilon(\mathbf{p}_v) \Rightarrow \mathbf{p}_u = \mathbf{p}_v, 
\label{eq:injectivity}
\end{equation}
or, more intuitively, that the environment and the sensor are rich enough so that the sensory state \( \chi_\epsilon(\mathbf{p}_u) = \mathbf{s}^\epsilon \) is characteristic of the sensor position \( \mathbf{p}_u \) without ambiguity.

Moreover, as mentioned in Section~\ref{sec:problem}, as displacements of the sensor are equivalent from a sensory perspective to an (opposite) displacement of the environment, we assume \emph{sensory compensability}:
\begin{equation}
\forall \epsilon, \forall \mathbf{p}, \forall \delta :
\chi_{\epsilon}(\mathbf{p} + \delta) = \chi_{\epsilon - \delta}(\mathbf{p}),
\label{eq:compensability}
\end{equation}
where $\epsilon - \delta$ denotes a change applied only to the spatial component(s) of $\epsilon$.

Given: \( (\mathbf{m}_{a'},\mathbf{m}_{b'}) \in \big[(\mathbf{m}_a,\mathbf{m}_b)\big]_{ff}\)
we have:
\begin{equation}
\begin{split}
\overrightarrow{f(\mathbf{m}_{a'})f(\mathbf{m'}_{b'})} &= \overrightarrow{f(\mathbf{m}_a)f(\mathbf{m}_b)}
\\
\Leftrightarrow  \overrightarrow{\mathbf{p}_{a'}\mathbf{p}_{b'}} &= \overrightarrow{\mathbf{p}_a \mathbf{p}_b}
\\
\Leftrightarrow \mathbf{p}_{b'} - \mathbf{p}_{a'} &= \mathbf{p}_{b} - \mathbf{p}_{a}.
\end{split}
\label{eq:parallel}
\end{equation}

Then, having \( \phi_{\epsilon'}(\mathbf{m}_{a'})= \phi_{\epsilon}(\mathbf{m}_{a}) \), we have:
\begin{equation}
\begin{split}
\forall \epsilon, \forall \epsilon': \phi_{\epsilon'}(\mathbf{m}_{a'}) &= \phi_{\epsilon}(\mathbf{m}_{a}) = \mathbf{s}_a^\epsilon
\\
\Leftrightarrow\: \chi_{\epsilon'}(\mathbf{p}_{a'}) &= \chi_{\epsilon}(\mathbf{p}_{a})
\\
\Leftrightarrow\: \chi_{\epsilon - \delta}(\mathbf{p}_{a'})
&= \chi_{\epsilon}(\mathbf{p}_{a})
\\
\eqref{eq:compensability}\;\; \Leftrightarrow\: \chi_{\epsilon}(\mathbf{p}_{a'} + \delta)
&= \chi_{\epsilon}(\mathbf{p}_{a})
\\
\eqref{eq:injectivity}\;\; \Leftrightarrow\: \mathbf{p}_{a'} + \delta &= \mathbf{p}_{a}
\end{split}
\label{eq:shift}
\end{equation}
and thus:
\begin{equation}
\begin{split}
\phi_{\epsilon'}(\mathbf{m}_{b'}) &= \chi_{\epsilon'}(\mathbf{p}_{b'}) \\
&= \chi_{\epsilon - \delta}(\mathbf{p}_{b'}) \\
\eqref{eq:compensability}\;\; &= \chi_{\epsilon}(\mathbf{p}_{b'} + \delta) \\
&= \chi_{\epsilon}(\mathbf{p}_{b'} + \delta + \mathbf{p}_{a'} - \mathbf{p}_{a'}) \\
\eqref{eq:shift}\;\; &= \chi_{\epsilon}(\mathbf{p}_{b'} + \mathbf{p}_{a} - \mathbf{p}_{a'}) \\
\eqref{eq:parallel}\;\; &= \chi_{\epsilon}(\mathbf{p}_{b} + \mathbf{p}_{a} - \mathbf{p}_{a}) \\
&= \chi_{\epsilon}(\mathbf{p}_{b}) = \mathbf{s}_{b}^{\epsilon}
\end{split}
\end{equation}

\end{document}